\documentclass[sigconf]{acmart}

\usepackage{soul}
\usepackage{colortbl}
\usepackage{color, xcolor}
\usepackage{natbib}
\usepackage{algorithm}
\usepackage{algorithmic}
\usepackage{multirow}
\usepackage{multicol}
\usepackage{amsmath}
\usepackage{booktabs} 
\usepackage{diagbox}
\usepackage{arydshln}
\usepackage[most]{tcolorbox}  

\usepackage{balance}
\usepackage{graphicx}
\usepackage{subcaption}
\usepackage{makecell}

\AtBeginDocument{%
  }

\setcopyright{acmlicensed}
\copyrightyear{2026}
\acmYear{2026}
\setcopyright{cc}
\setcctype{by}
\acmConference[MM '26] {Proceedings of the 35th ACM International Conference on Multimedia}{November 10--14, 2026}{Rio de Janeiro, Brazil.}
\acmBooktitle{Proceedings of the 35th ACM International Conference on Multimedia (MM '26), November 10--14, 2026, Rio de Janeiro, Brazil}
\acmISBN{979-8-4007-2213-4/2026/11}
\acmDOI{10.1145/3767308.3836426}
\begin{document}

\title{Let the Bullets Fly: Multimodal Fake News Detection with Temporal-Aligned Generative Danmaku}

\author{Xiansheng Luo}
\orcid{0009-0006-5019-4837}
\authornote{Equal Contribution}
\affiliation{%
  \institution{Yangzhou University}
  \city{Yangzhou}
  \state{Jiangsu}
  \country{China}}
\email{mz220240305@stu.yzu.edu.cn}

\author{Chaowei Zhang}
\orcid{0000-0001-5051-5318}
\authornotemark[1]
\authornote{Corresponding Author}
\affiliation{%
  \institution{Yangzhou University}
  \city{Yangzhou}
  \state{Jiangsu}
  \country{China}}
\email{cwzhang@yzu.edu.cn}
\renewcommand{\shortauthors}{Chaowei Zhang}

\author{Zewei Zhang}
\orcid{0009-0005-1716-4102}
\affiliation{%
  \institution{Auburn University}
  \city{Auburn}
  \state{Alabama}
  \country{USA}
}
\email{zez0001@auburn.edu}
\author{Yi Zhu}
\orcid{0000-0003-3045-2588}

\affiliation{%
  \institution{Yangzhou University}
  \city{Yangzhou}
  \state{Jiangsu}
  \country{China}}
\email{zhuyi@yzu.edu.cn}	
\renewcommand{\shortauthors}{Yi Zhu}

\author{Jipeng Qiang}
\orcid{0000-0001-5721-0293}
\affiliation{%
  \institution{Yangzhou University}
  \city{Yangzhou}
  \state{Jiangsu}
  \country{China}}
\email{jpqiang@yzu.edu.cn}

\renewcommand{\shortauthors}{Xiansheng Luo, Chaowei Zhang, Zewei Zhang, Yi Zhu, and Jipeng Qiang}

\begin{abstract}

The social interactions among crowds via \textit{Danmaku} (a.k.a., bullet comments) on modern multimedia platforms can facilitate both viewpoint conflicts and consensus, providing fine-grained discriminative social signals that can benefit fake news detection. However, the inherent accumulation latency of \textit{Danmaku} in real-world scenarios violates the real-time necessity of fake news detection, making the studies of \textit{Danmaku}-related fake news detection underexplored. To break this violation, we simulate this temporal-aware user interactive process by proposing a novel temporal \textbf{Gen}erative \textbf{da}nmaku framework, called \textbf{Genda}, which consists of: (1) a \textit{Danmaku} Trigger for predicting the timing and intensity of user reactions; and (2) a \textit{Danmaku} Generator for synthesizing corresponding semantic and emotional expressions, thereby mutually constructing a temporally aligned and human-like pseudo \textit{Danmaku} streams. To make the generated \textit{Danmaku} useful for identifying fake news videos, we further design a \textit{Danmaku}-guided Temporal Multimodal fake news detection model - \textbf{DM-FEND}, which enables fine-grained multimodal interactions among video, audio, text, and \textit{Danmaku},  enhancing dynamic modalities alignment and semantic noise inhibition. The experimental results demonstrate that \emph{DM-FEND} consistently outperforms state-of-the-art baselines across both Chinese (FakeSV) and English (FakeTT) benchmarks. Further ablations validate the crucial role of temporal \textit{Danmaku} modeling in enhancing robustness and discriminative capability. Finally, this study offers a bright and robust solution for multimodal fake news detection in modern social interactive fashions by bridging the temporal inconsistency between news and user behaviors. To ensure reproducibility, the code and data used in this study are released at: \url{https://github.com/126541/Let-the-Bullets-Fly}.

\end{abstract}

\begin{CCSXML}
<ccs2012>
   <concept>
       <concept_id>10002978.10003029</concept_id>
       <concept_desc>Security and privacy~Human and societal aspects of security and privacy</concept_desc>
       <concept_significance>500</concept_significance>
       </concept>
   <concept>
       <concept_id>10002951.10003227.10003251</concept_id>
       <concept_desc>Information systems~Multimedia information systems</concept_desc>
       <concept_significance>500</concept_significance>
       </concept>
 </ccs2012>
\end{CCSXML}

\ccsdesc[500]{Security and privacy~Human and societal aspects of security and privacy}
\ccsdesc[500]{Information systems~Multimedia information systems}

\keywords{Fake News Detection, Temporal Danmaku Generation, Multimodal Alignment, Temporal Awareness}

\begin{comment}
\begin{teaserfigure}
 \includegraphics[width=\textwidth]{sampleteaser}
  \caption{Seattle Mariners at Spring Training, 2010.}
  \Description{Enjoying the baseball game from the third-base
  seats. Ichiro Suzuki preparing to bat.}
  \label{fig:teaser}
\end{teaserfigure}

\received{20 February 2007}
\received[revised]{12 March 2009}
\received[accepted]{5 June 2009}

\end{comment}
\maketitle

\section{Introduction}

The proliferation of new-fashion multimedia platforms like TikTok shifted news exhibition into a highly compressed and fast-paced mode~\cite{haitao2024tiktok,cheng2024like,newman2022publishers}, aggravating the spread of fake news and necessitating urgent and real-time detection strategies~\cite{ruak2023impact,ty2025exploring}. These new-shape platforms foster diverse and dynamic user engagement, including traditional static user comments~\cite{huo2025live} and temporal-aware Danmaku~\cite{sun2024vico,chen2024hotvcom}, which is known as bullet comments overlaying the video playback. These interactions can boost viewpoint collisions among crowds, catalyzing the formation of collective consensus or stark conflicts. Such highly discriminative social interactive signals provide invaluable contextual cues for fake news detection. Unlike conventional comments that typically reflect a global, post-hoc evaluation of the entire video, Danmaku is intrinsically endowed with a temporal nature and synchronizes precisely with the video content~\cite{fatima2025semantic}. In particular, the fine-grained temporal sensitivity of Danmaku is crucial for identifying fleeting segments of fabricated information within the video stream, highlighting its unique potential and imperative value for research. Despite this, Danmaku use in multimodal fake news detection remains underexplored due to the conflict between its accumulation latency in real-world scenarios and real-time fake news detection.

Current mainstream short video-oriented fake news detection predominantly focuses on analyzing the intrinsic authenticity of the multimedia content, including forensic traces of video splicing and editing, audio manipulations designed to hijack audience emotions~\cite{bu2024fakingrecipe,zhang2019detecting,huh2018fighting,wang2025mfae}, or multi-view causal reasoning~\cite{chen2023causal,liu2025deconfounded,zhang2026turning} and debiasing techniques~\cite{gong2025unseen,zhu2022generalizing,liu2024out}, to uncover deceptive visual patterns. However, these content-centric detectors struggle to capture the complex semantic traces and societal context surrounding the news, which often fall short when sophisticated manipulations leave negligible visual artifacts or true videos are maliciously miscontextualized. Some of the other video fake news detectors utilize social interaction signals, including propagation traces among crowds~\cite{zhang2023computational,li2025learning,kim2018leveraging,patidar2025fakethreads} and user comments~\cite{ye2025fake,nan2025exploiting,tan2026emotion}, as auxiliary modalities to extract the public's stance and collective wisdom. Given the sparsity of interactions during the early stages of news propagation, pioneering comment-based methods have leveraged LLMs to synthesize high-quality pseudo-comments for fake news detection~\cite{zhang2026acting,yang2026cross,yi2025challenges,zhang2025llms}. Can VLLMs be used to synthesize high-fidelity Danmaku streams for short videos? Beyond their temporal and frame-aligned characteristics, variations in Danmaku density reflect content significance and the intensity of user opinion conflicts. Thus, generating Danmaku that accurately models both timing and diverse semantic reactions remains a major challenge.

\begin{figure*}[t]
    \centering
    \begin{subfigure}[t]{0.47\textwidth}
        \centering
        \includegraphics[height=3.5cm]{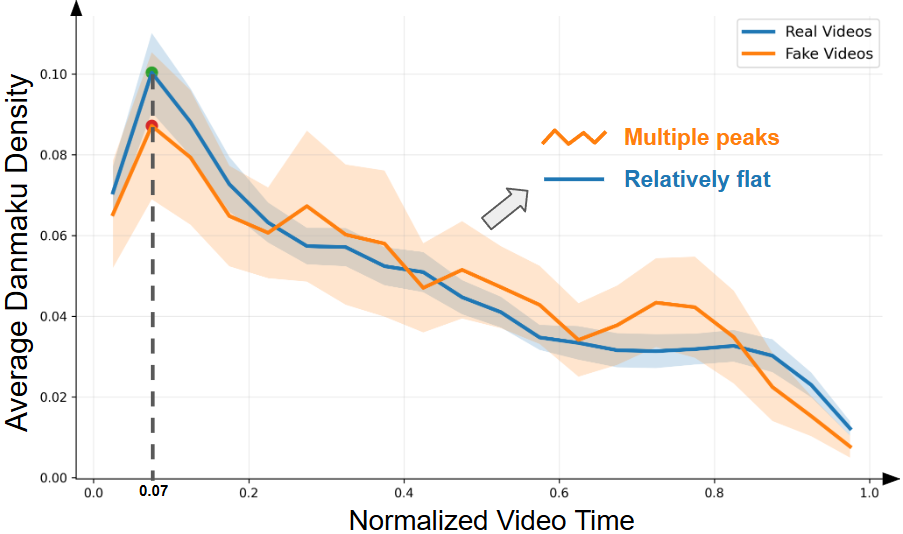}
        \caption{Temporal Density of Danmaku in Video}
        \label{fig:sub1}
    \end{subfigure}
    \hfill
    \begin{subfigure}[t]{0.5\textwidth}
        \centering
        \includegraphics[height=3.5cm]{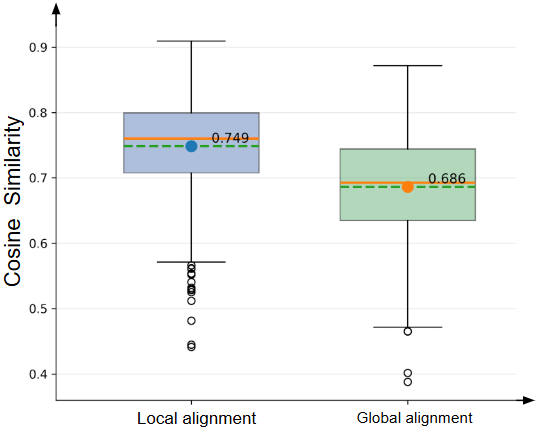}
        \caption{Semantic relatedness between \textit{Danmaku} and Video}
        \label{fig:sub2}
    \end{subfigure}
    \hfill
    \caption{%Empirical analysis of \textit{Danmaku} as a temporally structured and semantically grounded signal.
(a) Temporal distribution of \textit{Danmaku} across real and fake videos, showing consistent early-stage concentration and distinct evolution patterns.
(b) Semantic relatedness comparison between \textit{Danmaku} and video content.}
    \label{fig:Danmaku_analysis}
\end{figure*}

To encounter this obstacle, we propose \textbf{Genda}, a novel temporal Danmaku generation framework that simulates the time-aware interactive process of crowd responses. Unlike existing comment generation methods, \textit{Genda} models the underlying generation mechanism of user reactions during video playback, thereby reconstructing representative collective feedback in the absence of real-time interaction data. Specifically, \textit{Genda} decomposes the process of temporal-aligned Danmaku generation into two collaborative sub-tasks: (1) a Danmaku Trigger, which models the temporal evolution of video content and emotions to predict when user reactions are likely to occur and their corresponding intensity; and (2) a Danmaku Generator, which is conditioned on the trigger's signals, further generates semantic-aware and diverse emotional bullet comments aligned with the current video frames. Such synthetic temporally aligned Danmaku streams using \textit{Genda} compensate for the absence of user interaction signals in the early stage of news propagation, providing fine-grained supervision for subsequent modeling. Furthermore, the produced Danmaku can help the detectors locate potential anomalous segments and understand the evolving relationships among different modalities over time.

To take advantage of \emph{Genda} for facilitating fake news detection, we propose \textbf{DM-FEND}, a Danmaku-guided multimodal fake news detector, which leverages Danmaku as intermediate signals to facilitate the interactions among video, audio, and text at the segment level, thereby capturing local cross-modal inconsistencies and suppressing decision-irrelevant information. In detail, we first design a Danmaku-guided unimodal learning mechanism that reconstructs high-saliency tokens to strengthen decision-relevant semantic cues and suppresses low-saliency tokens to reduce semantic noise, thereby improving unimodal representation quality. Then, we further perform multimodal learning by modeling pairwise interactions among modalities and aligning them with Danmaku signals, aiming to capture cross-modal inconsistencies from the perspective of user reactions. Finally, \emph{DM-FEND} aggregates multimodal representations over the entire temporal sequence, jointly modeling content evolution and user responses to determine the authenticity of short videos. Overall, the main contributions of this study are presented as follows:
\begin{itemize}
    \item This study initially introduces Danmaku into multimodal fake news detection. To bridge the gap between Danmaku accumulation latency and the necessity of real-time detection, we propose a temporal Danmaku generation framework - \textit{Genda}, which consists of a Danmaku trigger and a Danmaku generator, aiming at constructing temporally aligned and human-like pseudo bullet comments. 

    \item To utilize the generated Danmaku stream for facilitating fake news detection, we develop \textit{DM-FEND}, which enables fine-grained multimodal interactions among video, audio, text, and Danmaku, enhancing dynamic modality alignment and semantic noise suppression.

    \item Our conducted extensive experiments on two short-video datasets demonstrate that \textit{DM-FEND} consistently outperforms baseline methods across various evaluation metrics, including fine-tuned unimodal models, prompt-based approaches on large language models, and SOTA baselines.
\end{itemize}

\section{Fake News Detection in the Era of LMs}

Mainstream studies on fake news detection primarily focus on analyzing the authenticity of news content across modalities, aiming to identify misleading patterns~\cite{rama2025dual,tsang2026misinformation}, semantic inconsistencies~\cite{yu2025sr}, or manipulation cues directly~\cite{wang2025fakesv} from the input data. In other words, these methods attempt to improve robustness and capture richer semantic relationships by integrating complementary signals across modalities, thereby assisting the task of fake news detection.  With the recent advancement of AI techniques, such a type of the approaches has evolved from traditional neural architectures to large-scale pre-trained models (a.k.a., LMs), and further to multimodal frameworks that jointly model heterogeneous information sources~\cite{li2025learning}. For example, recent relevant studies often leverage large language models or vision-language models to enhance reasoning ability and cross-modal understanding~\cite{wang2024llm,sun2024exploring,xie2024multiknowledge,zhang2025llms,hu2025synergizing}. However, these approaches heavily rely on content representations and implicit reasoning processes, overlooking the dynamic nature of information evolution. Specifically, they fail to explicitly model how users perceive, interpret, and react to content over time, leading to a gap between content understanding and user-level perception modeling.

Some of the other SOTA studies explore the usability of data augmentation~\cite{hua2023multimodal,hamed2025data,arik2026llm} and generation-based strategies~\cite{zhang2024mitigating,wang2024llm,peng2024automatic} in the area of fake news detection, aiming at improving the generalizability of detectors as well as enriching the diversity of news. Such types of approaches typically leverage LMs to generate supplementary supervision signals, such as reasoning chains~\cite{xu2024multimodal}, counterfactual samples~\cite{wang2025fakesv}, or extra multimodal content~\cite{fu2023multimodal}, thereby improving the diversity and informativeness of news data. For example, recent researchers manipulate prompting or instruction tuning to perform multi-step reasoning and uncover latent deceptive patterns~\cite{han2026beyond}, while some other generative approaches synthesize additional data distributions to improve robustness under domain shifts. Furthermore, a few studies also attempt to construct explanation-guided or reasoning-enhanced representations to reinforce the interpretability of fake news detection models~\cite{bai2024large}. However, they neither capture the temporal dynamics of user interactions nor learn how collective responses evolve alongside content, limiting their abilities in reflecting real-world perception processes.

Compared to the existing approaches, we shift our focus to temporally grounded user interaction modeling. Specifically, we propose a temporal Danmaku generation framework, called \textit{Genda}, which simulates the time-aware evolution of crowd interactions in news videos to produce high-quality pseudo Danmaku streams of news, thereby addressing the inherent latency of Danmaku in real-world scenarios as well as enabling the reconstruction of realistic interaction signals in early stages. Upon this, we further develop \textit{DM-FEND}, a Danmaku-guided multimodal framework that integrates generative Danmaku with video, audio, and text at the video clip level, thereby bridging the gap between content representation and user perception. Such a design allows our approach to capture clip-level cross-modal inconsistencies and reflect user temporal interaction in real-world news scenarios, leading to more robust and accurate fake news detection.

\section{Why does Danmaku work?}

To validate the usability of temporal-aware Danmaku streams in fake news detection, we collect 100 short news videos from Bilibili to examine the temporal distribution of Danmaku for real and fake videos. As shown in Fig.~\ref{fig:sub1}, we first compute the average Danmaku density over time by normalizing video timelines to $[0,1]$. It can be observed from Fig.~\ref{fig:sub1} that both real and fake videos exhibit a similar global pattern, where Danmaku density rapidly increases at the beginning (peaking happens at $t \approx 0.07$) and gradually declines thereafter, suggesting that Danmaku reflects structured collective responses. However, we can also observe a notable difference between real and fake videos, where real cases show a smooth and monotonic decay after the peak, yet fake ones exhibit stronger local fluctuations. Moreover, fake videos present larger variance across samples, indicating more diverse reaction patterns, whereas real videos demonstrate more consistent temporal dynamics.

To further examine the alignment between Danmaku and video content, we conduct the empirical study of  (1) the semantic relatedness comparison between Danmaku and corresponding local segments (See \emph{Local alignment} in Fig.~\ref{fig:sub2}), and (2) the the semantic relatedness comparison between Danmaku and the entire video (See \emph{Global alignment} in Fig.~\ref{fig:sub2}). As shown in Fig.~\ref{fig:sub2}, \emph{Local alignment} achieves a higher average semantic similarity (i.e., $0.749$) than that of \emph{Global alignment} (i.e., $0.686$), demonstrating that Danmaku is more closely associated with local semantics than global content. These observations suggest that Danmaku streams can be used as fine-grained temporal signals with both structured evolution and strong local alignment. Therefore, we can confirm that it is reasonable and essential to utilize Danmaku streams as dynamic social interactive signals for capturing temporal multimodal interactions, thereby facilitating the detection of fake news in terms of short videos. In the subsequent sections of the paper, we illustrate the details of our proposed temporal-aligned generative Danmaku framework - \emph{Genda} and Danmaku-guided multimodal fake news detection system - \emph{DM-FEND}.

\begin{figure*}[h!]
    \centering
    \includegraphics[width=0.96\linewidth,height = 0.5\linewidth]{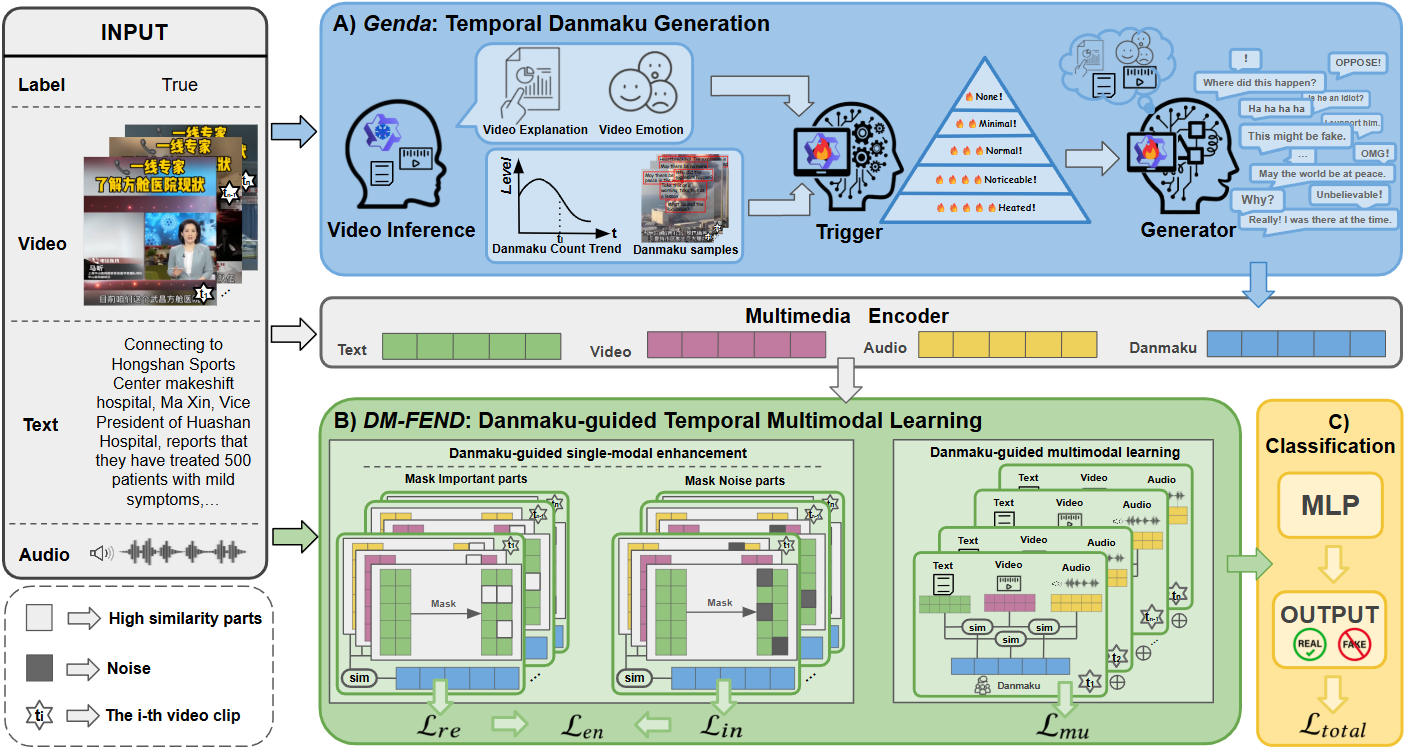}
    \caption{The workflow of our proposed approach. (A) \textit{Genda}: a temporal Generative Danmaku module that models user interaction dynamics by performing segment-level video understanding, followed by a Danmaku Trigger to estimate reaction intensity and a Danmaku Generator to produce semantically and temporally aligned Danmaku. (B) \textit{DM-FEND}: a Danmaku-guided temporal multimodal learning module that enhances unimodal representations via importance-aware and noise-aware masking, and captures fine-grained cross-modal interactions among text, video, audio, and Danmaku. (C) Classification block: temporally aggregated multimodal representations are fed into an MLP for final fake news prediction.}
    \label{fig:Jury_Selection}
\end{figure*}

\section{\textit{Genda}: Temporal Generative Danmaku}

The proposed \textit{Genda} framework simulates the temporal generation process of Danmaku by modeling how users perceive and react to video content over time. As illustrated in Fig.~\ref{fig:Jury_Selection}(A), there are three key modules involved in \textit{Genda}, including (1) \emph{video understanding module}, (2) \emph{Danmaku trigger}, and \emph{Danmaku generator}. More details about the implementation of \textit{Genda} are shown as follows.

\subsection{Problem Definition}

The fake news detection in short videos can be formulated as a binary classification task. Let $\mathcal{D} = \{V_1, V_2, \dots, V_n\}$ denote a dataset of short videos, where $n$ is the total number of news video samples. A randomly selected video $V_i$ is associated with a corresponding label $y_i \in \{0,1\}$, where $y_i=1$ indicates fake news, vice versa for $y_i=0$.
Specifically, $V_i$ maintains multiple modalities, including video, audio, and text. We denote the multimodal input as:
$V_i = \{x_i^{v}, x_i^{a}, x_i^{t}\}$,
where $x_i^{v}$, $x_i^{a}$, and $x_i^{t}$ represent the video, audio, and textual content, respectively.
To capture fine-grained temporal dynamics, each video is further decomposed into a sequence of temporal segments,
$V = \{ s_1, s_2, \dots, s_N \}$,
where $N$ is the number of segments in video $V$. For each segment $s_i$, we aim to model its corresponding Danmaku signal $d_{i,j}$, which reflects user reactions aligned with the temporal progression of the video. Since real Danmaku may not always be available in early stages, we employ a generation framework to approximate the Danmaku distribution:
$d_{i,j} \sim P_{\theta}(d \mid s_{i,j})$,
where $P_{\theta}$ denotes the learned generation model.
Based on the multimodal content and generated Danmaku, the objective is to learn a nonlinear mapping
$f: \{V_i, D_i\} \rightarrow y_i$,
where $D_i = \{d_{i,1}, d_{i,2}, \dots, d_{i,N_i}\}$ denotes the Danmaku sequence associated with video $V_i$. The goal is to leverage temporally aligned Danmaku as dynamic supervision to improve fake news detection.

\subsection{Video Understanding}

We decompose each video into a series of short segments and extract a structured representation of each segment to obtain fine-grained temporal semantics. Given an input video $V$ with duration $T$, we partition it into $N$ non-overlapping segments, as shown in the formula below~\ref{eq:semporal_segmentation}. %\textbf{Temporal Segmentation and Representation.} 
\begin{equation}
V = \{ s_1, s_2, \dots, s_N \}, \quad N = \left\lfloor \frac{T}{\Delta t} \right\rfloor + 1,
\label{eq:semporal_segmentation}
\end{equation}
where each segment has a fixed duration $\Delta t = 4$ seconds. For each segment $s_i$, we uniformly sample 12 frames $\mathcal{F}_i$ to preserve both static and dynamic visual cues, ensuring that temporal transitions and salient events are adequately captured. Then, we employ a vision-language model $\mathcal{M}_{vl}$ to perform segment-level multimodal reasoning. Each segment is represented as: %\textbf{Multimodal Understanding with Temporal Awareness.}
\begin{equation}
\begin{aligned}
z_i = \mathcal{M}_{vl}(\mathcal{F}_i, \text{text}, p_i) =\{c_i, e_i, p_i\}
\end{aligned}
\end{equation}
where $c_i$ and $e_i$ denote semantic content and emotional tone, $p_i = \frac{i}{N}$ encodes the relative temporal position. By injecting temporal context, the model learns to interpret each segment within the global narrative, capturing its semantic content, emotional tone, narrative role, and potential reaction cues. The video modality is finally represented as a temporally ordered sequence:
\begin{equation}
Z = \{ z_1, z_2, \dots, z_N \},
\end{equation}
where $z_i$ can be viewed as a \textit{semantic trajectory} at $i$th clip stamp of $Z$ that jointly encodes corresponding content evolution, emotional dynamics, and interaction signals. This representation provides a unified foundation for modeling temporally aligned user reactions in subsequent modules.

\subsection{Danmaku Trigger}

The Danmaku Trigger aims to model \textit{when} and \textit{how strongly} users react to video content by learning a temporal reaction intensity function over video segments. Unlike conventional classification-based approaches, we formulate this problem as an ordinal-aware latent intensity modeling task, which captures the gradual evolution of user engagement. For each segment $s_i$, we construct a unified representation by integrating semantic, emotional, and contextual signals derived from the Video Understanding module, as shown in the formula below~\ref{eq:danmaku_trigger1}.
\begin{equation}
h_i = f_{\theta}(z_i, m_i, h_i^{meta})
\label{eq:danmaku_trigger1}
\end{equation}
where  $m_i$ and $h_i^{meta}$ represent auxiliary contextual information such as textual hints and video popularity. The function $f_{\theta}(\cdot)$ is instantiated by a large language model with LoRA adaptation, enabling efficient yet expressive modeling of temporal user perception. For modeling ordinal intensity, we quantify user reaction as a latent continuous intensity score $\varphi_i$, which reflects the underlying strength of audience engagement for segment $s_i$. Instead of directly predicting discrete labels, we project this continuous score onto ordinal levels through a set of learnable monotonic thresholds:
\begin{equation}
P(y_i > k \mid \varphi_i) = \sigma\!\left(\varphi_i - \sum_{j=1}^{k} \text{softplus}(\delta_j)\right),
\end{equation}
where $y_i \in \{0,1,2,3,4\}$ denotes the discrete Danmaku intensity level of segment $s_i$, corresponding to \textit{None}, \textit{Minimal}, \textit{Normal}, \textit{Noticeable}, and \textit{Heated}, respectively. The variable $k \in \{0,1,2,3\}$ represents the ordinal threshold index, and the condition $y_i > k$ indicates whether the reaction intensity exceeds the $k$-th level.
Here, $\{\delta_j\}$ are unconstrained parameters used to construct ordered thresholds via cumulative softplus, ensuring monotonicity, and $\sigma(\cdot)$ denotes the sigmoid function.

During inference, the predicted intensity level $\hat{y}_i$ is determined by counting how many thresholds are surpassed by $\varphi_i$, resulting in a final prediction in the same ordinal set $\{0,1,2,3,4\}$.
This formulation captures the ordinal structure of reaction intensity while allowing smooth transitions across levels, enabling the model to represent subtle variations in user engagement as a continuous latent process.

\subsection{Danmaku Generator}

The Danmaku Generator aims to synthesize human-like Danmaku conditioned on video content and predicted user reaction signals. Building upon the segment representation $z_i$ (Video Understanding) and the reaction intensity $\hat{y}_i$ (Danmaku Trigger), we formulate the generation process as a conditional social response modeling task. For each segment $s_i$, we generate Danmaku conditioned on semantic content, predicted reaction intensity, temporal context, and latent user style. Specifically, the generation process is modeled as:
\begin{equation}
d_{i,j} \sim P_{\theta}\big(d \mid z_i, \hat{y}_i, p_i, s_j\big), \quad s_j \sim \mathcal{S},
\end{equation}
where $z_i$ encodes semantic and emotional information, $\hat{y}_i$ denotes the predicted reaction intensity level, $p_i$ represents the temporal position, and $s_j$ is a latent style variable sampled from a predefined distribution $\mathcal{S}$ to simulate diverse user personas.
Under this formulation, $\hat{y}_i$ controls the overall reaction strength, determining both the content characteristics and the number of generated Danmakus, while $s_j$ introduces stylistic variability to produce natural and diverse expressions. This design enables the generator to produce temporally aligned, semantically coherent, and human-like Danmaku, effectively approximating real user interaction behaviors.

We implement the generator using a large language model with LoRA fine-tuning, enabling efficient adaptation while preserving strong generative capabilities. As a result, the generated Danmaku forms a temporally coherent and human-like interaction stream, where each segment is associated with responses that reflect its semantic content, emotional tone, and reaction intensity. Furthermore, we analyzed Pseudo Danmaku Quality in section A of the supplementary document.

\section{Danmaku-guided Fake News Detection}
Our proposed Danmaku-guided temporal multimodal learning framework, \textit{DM-FEND}, effectively leverages temporally aligned Danmaku signals for fake news detection.  As illustrated in Fig.~\ref{fig:Jury_Selection}, the framework first employs a \textbf{Multimedia Encoder} to extract segment-level representations from multiple modalities, including text, video, audio, and Danmaku. Specifically, each modality is encoded into a sequence of feature embeddings, forming a unified multimodal representation space that preserves both semantic and temporal information.
Based on these encoded features, \textit{DM-FEND} further consists of two complementary components: (1) Danmaku-guided unimodal learning, which refines each modality under the guidance of Danmaku signals, and (2) Danmaku-guided multimodal learning, which models cross-modal interactions and aligns them with user reaction dynamics, as illustrated in Fig.~\ref{fig:Jury_Selection}(B)

\subsection{Danmaku-guided Unimodal Learning}

To enhance unimodal representations under the guidance of Danmaku, we design a Danmaku-conditioned masking and reconstruction framework that enables the model to focus on salient content while suppressing noise.
For a given video segment $s_i$, let $d_i$ denote the Danmaku representation generated by the Danmaku Generator, and let $x_i^m \in \mathbb{R}^{L_m \times D}$ denote the token-level feature sequence of modality $m \in \{\text{video}, \text{audio}\}$, where $L_m$ is the number of tokens and $D$ is the feature dimension.
We first compute a Danmaku-conditioned saliency score:
\begin{equation}
\alpha_i = \mathcal{S}(x_i^m, d_i), \quad \alpha_i \in \mathbb{R}^{L_m},
\end{equation}
where $\mathcal{S}(\cdot)$ maps each token to a scalar importance value, reflecting how strongly it is associated with user reactions encoded in $d_i$.
Based on $\alpha_i$, we derive two complementary token subsets:
- $M_i^{top}$: the index set of top-ranked tokens (high-saliency),
- $M_i^{low}$: the index set of bottom-ranked tokens (low-saliency).
We then construct masked representations by replacing selected tokens with a learnable mask embedding and re-encoding them. Let $\hat{x}_i^m$ denote the reconstructed features from $M_i^{top}$ masking, and let $\tilde{x}_i^m$ denote the features under $M_i^{low}$ masking. The overall training objective is defined as shown in Eq.~\ref{eq:DM-FEND1}.
\begin{equation}
\scalebox{0.88}{$\mathcal{L}_{en} =
\overbrace{\frac{1}{|M_i^{top}|} \sum_{t \in M_i^{top}} \| \hat{x}_{i,t}^m - x_{i,t}^m \|_2^2}^{\mathcal{L}_{re}}
+
\overbrace{1 - \cos\big(\text{Pool}(x_i^m), \text{Pool}(\tilde{x}_i^m)\big)}^{\mathcal{L}_{in}} 
+
\mathcal{R}(s_i),$}
\label{eq:DM-FEND1}
\end{equation}
where:
- $\hat{x}_{i,t}^m$ denotes the reconstructed feature of token $t$,
- $\text{Pool}(\cdot)$ is a mean pooling function over tokens,
- $\cos(\cdot,\cdot)$ denotes cosine similarity,
- $\mathcal{R}(s_i)$ is a regularization term encouraging a balanced saliency distribution. The first term $\mathcal{L}_{re}$ enforces the model to recover masked high-saliency tokens, encouraging it to capture semantically important content. The second term $\mathcal{L}_{in}$ promotes invariance by aligning the original representation with the noise-masked representation, improving robustness to irrelevant information. The third term $\mathcal{R}(s_i)$ prevents degenerate solutions by regularizing the saliency distribution.
Through this dual masking mechanism, the model learns to distinguish informative content from noise under the guidance of Danmaku $d_i$, leading to more robust and discriminative unimodal representations.

\begin{table*}[htbp!]
\renewcommand{\arraystretch}{1}
\centering
\caption{This paper validates the performance of various baseline methods against our proposed method on the FakeSV and FakeTT datasets. Best results are highlighted in \textbf{bold}, and second-best results are highlighted with \underline{underlines}. Paired t-tests were used to test the statistical significance of all baselines (p < 0.05), and significant differences are indicated by (*).}
\label{tab:video_comparison}
\resizebox{\linewidth}{!}{
\begin{tabular}{@{}lcccccccc@{}}
\toprule
\textbf{\multirow{2}{*}{\large{Models}}} & \multicolumn{4}{c}{\textbf{FakeSV}} & \multicolumn{4}{c}{\textbf{FakeTT}} \\ 
\cmidrule(lr){2-5} \cmidrule(lr){6-9}
& \textbf{Acc} & \textbf{Mac-F1} & \textbf{Prec}  & \textbf{Rec} & \textbf{Acc} & \textbf{Mac-F1} & \textbf{Prec} & \textbf{Rec}  \\ \midrule
\rowcolor{gray!20}  \multicolumn{9}{c}{\textit{Unimodal Baselines}} \\
 Text(Bert) & 81.36 & 81.29 & 81.36   & 81.33 & 76.54 & 75.00  &74.70 &77.24   \\ 
 Image(CLIP-vit) & 73.99 & 73.97 & 75.65   & 75.35 &68.90 & 68.49 &71.02 &73.73     \\ 
 Audio(wav2vec2) & 75.52 & 75.13  & 76.86 & 76.37 & 70.54  &69.77 &71.75 & 74.00  \\ 
 Video(VideoMAEv2) & 72.86 & 72.58 & 72.53  & 72.71 & 73.24 & 71.92  &71.81 &74.39  \\ \hdashline
\rowcolor{gray!20}
\multicolumn{9}{c}{\textit{LMs Prompting Benchmarks}} \\
GPT-5-mini & 75.46 & 74.04 & 76.54 & 73.70 & 64.49  &56.74 &66.00 & 58.97   \\ 
InternVL2.5-8B &67.51 & 66.87 & 67.85  & 66.38 & 59.27 &  58.33  &60.25 &59.64  \\ 
Qwen2.5-VL-7B & 66.19 & 65.60 & 67.39  & 67.22 & 59.09 & 58.37  &59.74 &59.38 \\ \hdashline
\rowcolor{gray!20}  \multicolumn{9}{c}{\textit{Multimodal Detectors}} \\
 SV-FEND~(\small\citeauthor{qi2023fakesv} 2023) & 80.88 & 80.54  & 80.18 & 80.62 & 77.14  &75.63 &75.12 &77.56\\ 
FakingRecipe~(\small\citeauthor{bu2024fakingrecipe} 2024) & 85.06 & 84.54 & 85.69  & 84.08 & 79.60 & 77.94  &77.25 &79.39 \\
ExMRD~(\small\citeauthor{hong2025following} 2025) & 83.39 & 82.81 & 83.97  & 82.37 & \underline{79.80} & \underline{79.19}  &\underline{78.68} &\underline{81.93} \\
FakeSV-VLM~(\small\citeauthor{wang2025fakesv} 2025) & \underline{86.16} & \underline{85.74} & \underline{86.61}  & \underline{85.34} & 78.93 & 77.85  &77.47 &80.68\\
SGAN~(\small\citeauthor{li2026anchor} 2026) & 85.24 & 84.88 & 85.35  & 84.61 & 79.60 & 77.76  &77.12 &78.88 \\
\hdashline

\textbf{\emph{DM-FEND}} & \textbf{87.01*} & \textbf{86.68*} & \textbf{86.77*}  & \textbf{90.66*} & \textbf{83.43*} & \textbf{82.92*}  &\textbf{85.94*} &\textbf{85.80*} \\ 
\bottomrule 
\end{tabular}}
\end{table*}

\subsection{Danmaku-guided Multimodal Learning}

Based on the enhanced unimodal representations, we further model cross-modal interactions under the guidance of temporally aligned Danmaku. The key idea is to treat Danmaku $d_i$ as an intermediate supervisory signal that aligns different modalities at the segment level and enhances discriminative representation learning. For each segment $s_i$, let $x_i^{m} \in \mathbb{R}^{D}$ denote the enhanced representation of modality $m \in \{\text{text}, \text{video}, \text{audio}\}$, and let $d_i \in \mathbb{R}^{D}$ denote the corresponding Danmaku representation. We construct pairwise multimodal representations and align them with Danmaku via contrastive learning. The crossmodal consistency loss is defined as:
\begin{equation}
\mathcal{L}_{cm}
=
\frac{1}{3}
\sum_{(a,b)}
\mathcal{L}_{co}\!\left(d_i, \phi(x_i^{a}, x_i^{b})\right),
\end{equation}
where $(a,b) \in \{(\text{text},\text{video}), (\text{text},\text{audio}), (\text{video},\text{audio})\}$, $\phi(\cdot,\cdot)$ denotes a fusion function for modality pairs, and $\mathcal{L}_{co}(\cdot,\cdot)$ is an InfoNCE-based contrastive loss. This objective encourages modality pairs that correspond to the same segment to be aligned with the associated Danmaku, while pushing apart mismatched pairs.
To acquire the representation of videos, we aggregate segment-level multimodal representations across time to obtain a video-level feature, and optimize it jointly with the classification and similarity objectives:
\begin{equation}
\left\{
\begin{aligned}
h_i &= \mathcal{A}\big(\{x_i^{m}\}_{m}, \{d_i\}\big), \\
\mathcal{L}_{mu} &= \mathcal{L}_{cl} +  \mathcal{L}_{cm} +  \mathcal{L}_{vs},
\end{aligned}
\right.
\end{equation}
where $\mathcal{A}(\cdot)$ denotes temporal aggregation and multimodal fusion, $h_i$ is the video-level representation, $\mathcal{L}_{cl}$ is the classification loss, $\mathcal{L}_{cm}$ is the Danmaku-guided cross-modal consistency loss.
Overall, this module leverages Danmaku as a temporally aligned supervision signal to guide both segment-level cross-modal alignment and video-level representation learning, enabling the model to capture consistent patterns as well as subtle cross-modal inconsistencies for fake news detection.

\subsection{Binary Classification}
Based on the learned multimodal representation, we perform final fake news classification at the video level. Specifically, the aggregated representation $h_i \in \mathbb{R}^{D}$ encodes both multimodal content evolution and user reaction dynamics, as illustrated in Fig.~\ref{fig:Jury_Selection}(C)
The entire framework is optimized in an end-to-end manner with a unified objective:
\begin{equation}
\mathcal{L}_{total}
=
\lambda_1 \mathcal{L}_{en}
+
\lambda_2 \mathcal{L}_{mu},
\end{equation}
$\lambda_1, \lambda_2$ are trade-off hyperparameters.
This unified formulation enables joint optimization of representation learning, cross-modal alignment, and final decision-making under Danmaku supervision.

\section{Experiments}

\textbf{\Large{Implementation Details:}} We use \textit{Qwen2.5-VL-7B-Instruct\footnote{https://huggingface.co/Qwen/Qwen2.5-VL-7B-Instruct}} for segment-level video understanding and LoRA-tune \textit{Qwen2.5-7B-Instruct\footnote{https://huggingface.co/Qwen/Qwen2.5-7B-Instruct}} as both the Danmaku Trigger and Generator. Training uses a maximum sequence length of 896, batch size 2, gradient accumulation 8, learning rate $4\times10^{-5}$, 2 epochs, and FP16 precision. We define five ordinal classes—\textit{None}, \textit{Minimal}, \textit{Normal}, \textit{Noticeable}, and \textit{Heated}—with four thresholds and a distribution regularization coefficient of 0.2.
For \textbf{\textit{DM-FEND}}, we train for 30 epochs using AdamW with batch size 4, a base learning rate of $3\times10^{-5}$, a text encoder learning rate of $1\times10^{-6}$, and weight decay of $5\times10^{-4}$. Audio and video feature dimensions are 768 and 1408, respectively. The model uses a hidden size of 768, 8 attention heads, 2 temporal layers, and a dropout rate of 0.1.
Performance is evaluated using Accuracy (Acc), Macro F1 (Mac-F1), fake-class F1 (Mis-F1), Precision (Prec), and Recall (Rec), as shown in Tables\ref{tab:video_comparison} \& \ref{tab:ablation}.\\ \\
\textbf{\Large{Datasets}:} We evaluate our approach on two real-world fake news video benchmarks, FakeSV~\cite{qi2023fakesv}\footnote{https://github.com/ICTMCG/FakeSV} and FakeTT~\cite{bu2024fakingrecipe}\footnote{https://github.com/ICTMCG/FakingRecipe}.\\\\ %Consistent with previous work~\cite{bu2024fakingrecipe}, the datasets are temporally split based on video release time into training, validation, and test sets with a ratio of 70\%, 15\%, and 15\%, respectively.\\ \\
\textbf{\Large{Baselines}:} We select diverse baselines for performance comparison, including unimodal baselines, LMs prompting benchmarks, and SOTA multimodal fake news detectors. For unimodal baselines, we evaluate text, image, audio, and video models based on BERT~\cite{devlin2019bert}, CLIP-ViT~\cite{radford2021learning}, wav2vec2~\cite{baevski2020wav2vec}, and VideoMAEv2~\cite{tong2022videomae}, respectively. For LMs benchmarks, we adopt GPT-5-mini~\cite{singh2025openai}, InternVL2.5-8B~\cite{chen2024internvl}, and Qwen2.5-VL-7B~\cite{peng2023yarn}. 
For multimodal detectors, we investigate various representative approaches, including SV-FEND~\cite{qi2023fakesv}, FakingRecipe~\cite{bu2024fakingrecipe}, ExMRD~\cite{hong2025following}, FakeSV-VLM~\cite{wang2025fakesv}, and SGAN~\cite{li2026anchor}. These baselines cover a wide range of paradigms, enabling a comprehensive evaluation of our proposed \emph{DM-FEND}.

\subsection{Results and Analysis}
Table~\ref{tab:video_comparison} presents the performance comparison between our proposed \emph{DM-FEND} and various baseline methods on the FakeSV and FakeTT datasets. From the results, we draw the following observations.

First, \emph{DM-FEND} consistently achieves the best performance across all evaluation metrics on both datasets. On FakeSV, our method attains 87.01\% in accuracy and 86.68\% in Macro-F1, outperforming the strongest baseline, FakeSV-VLM, by 0.85\% and 0.94\%, respectively. More notably, our model achieves a substantial improvement in Recall, indicating a stronger ability to detect difficult or ambiguous fake news instances. On FakeTT, the improvements are even more significant, where \emph{DM-FEND} surpasses the best baseline (ExMRD) by 3.63\% in accuracy and 3.73\% in Macro-F1. These gains validate the effectiveness of temporally aligned Danmaku.

Second, unimodal models and large language models (LMs) exhibit relatively limited performance compared to multimodal approaches. For example, the best unimodal model (Text-BERT) achieves 81.36\% accuracy on FakeSV and 76.54\% on FakeTT, which is significantly lower than multimodal methods. Similarly, LMs such as GPT-5-mini achieve only 64.49\% accuracy on FakeTT, indicating that relying solely on reasoning without explicit multimodal alignment is insufficient. These results highlight the necessity of modeling cross-modal interactions for fake news detection.

Third, compared with state-of-the-art multimodal methods, \emph{DM-FEND} demonstrates clear advantages. While strong baselines such as FakingRecipe and SGAN achieve competitive results (e.g., around 85\% accuracy on FakeSV and 79\% on FakeTT), they mainly rely on global fusion or static alignment. In contrast, our method explicitly models temporally aligned Danmaku as dynamic supervision, enabling fine-grained segment-level alignment and noise suppression, which leads to more robust performance. Finally, we conduct further ablation studies and detailed analyses in subsequent sections to investigate the contributions of each component.

\subsection{Ablation Study}

\begin{table}[ht!]
\centering

\caption{Ablation study of different components. DA: Temporal Danmaku Generation, EN: Danmaku-guided unimodal learning
, CM: Danmaku-guided multimodal learning.}
\label{tab:ablation}
\renewcommand{\arraystretch}{1}
\resizebox{\linewidth}{!}{
\begin{tabular}{c c c  cccc}
\toprule
\multirow{2}{*}{\textbf{DA}} & \multirow{2}{*}{\textbf{EN}} & \multirow{2}{*}{\textbf{CM}}
& \multicolumn{2}{c}{\textbf{FakeSV}} 
& \multicolumn{2}{c}{\textbf{FakeTT}} \\
\cmidrule(lr){4-5} \cmidrule(lr){6-7}
 &  &  
& \textbf{ Acc} & \textbf{Mis-F1}  
& \textbf{Acc} & \textbf{Mis-F1 } \\
\midrule

$\times$ & $\times$ &  $\times$
& 81.49 & 80.64  
& 77.58 & 76.95  \\
\hdashline

$\times$ & $\times$ &  
& 83.12 & 82.87  
& 79.79 & 79.81  \\

$\times$ &  &  $\times$
& 83.72 & 82.47  
& 80.39 & 80.26  \\

 & $\times$ & $\times$ 
& 82.95 & 84.83  
& 80.16 & 83.42  \\
\hdashline

$\times$ &  & 
& 84.32 & 83.56   
& 81.18 & 80.20  \\

 & $\times$ &  
& 83.39 & 85.58  
& 80.64 & 83.66  \\

 &  &  $\times$
& \underline{84.39} & \underline{86.34} 
& \underline{82.09} & \underline{84.46}  \\

\hdashline
\multicolumn{3}{c}{\textbf{\emph{DM-FEND}}}  
& \textbf{87.01} & \textbf{88.64}  
& \textbf{83.43} & \textbf{85.84}  \\

\bottomrule
\end{tabular}}
\end{table}

\begin{figure}[ht]
    \centering
    \includegraphics[width=1\linewidth,height = 0.55\linewidth]{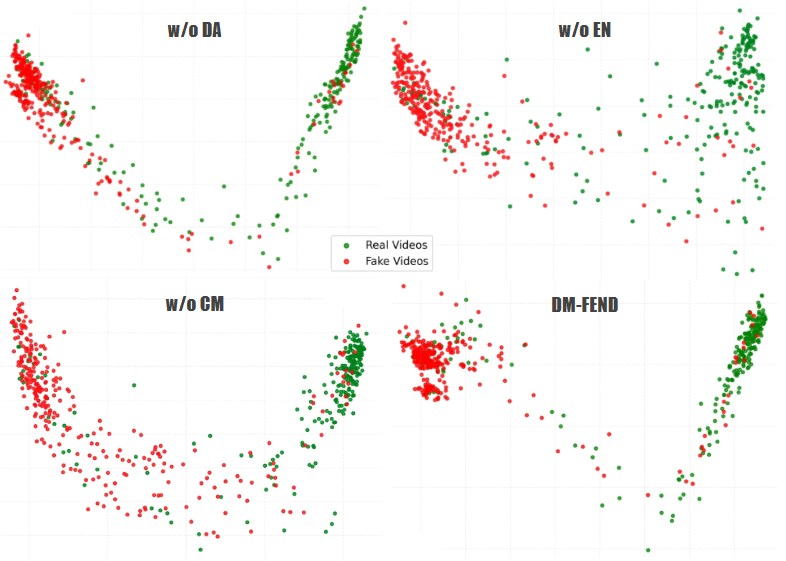}
    \caption{Visualized comparison between DM-FEND and its representative variants on the FakeSV dataset using Principal Component Analysis (PCA). Green represents real news sample points, and red represents fake news sample points.}
    \label{fig:abliation_img}
\end{figure}

To investigate the contribution of each component in our framework, we conduct ablation studies on three key modules: temporal Danmaku generation (DA), Danmaku-guided unimodal learning (EN), and Danmaku-guided multimodal learning (CM). The results are summarized in Table~\ref{tab:ablation}, and the corresponding feature distributions are visualized in Fig.~\ref{fig:abliation_img}.

First, when none of the proposed components are applied, the model achieves an accuracy of 81.49\% on FakeSV and 77.58\% on FakeTT, showing relatively low performance. This indicates that standard multimodal modeling without Danmaku guidance is insufficient to capture complex cross-modal inconsistencies.
Second, introducing each module individually consistently improves performance. Specifically, incorporating only the CM module increases the accuracy on FakeSV by 1.63\%, demonstrating the effectiveness of modeling cross-modal interactions. Similarly, adding only EN further improves performance, indicating that refining unimodal representations helps reduce noise and improve feature quality. When only DA is introduced, the model also achieves notable gains, with the Mis-F1 on the FakeTT dataset improving by 6.47\%, showing that Danmaku provides effective supervisory signals.
Third, combining multiple modules further enhances performance. Notably, even without the Danmaku generation module (DA), jointly applying EN and CM already achieves strong results, reaching 84.32\% accuracy on FakeSV. This indicates that enhancing unimodal representations and modeling cross-modal interactions alone can provide substantial improvements.
Further incorporating Danmaku with other modules leads to additional gains. For example, integrating DA with CM improves the accuracy on FakeSV to 83.39\%, while combining DA with EN achieves 84.39\%. These results indicate that Danmaku plays a complementary role by enhancing both unimodal optimization and cross-modal alignment. From Fig.~\ref{fig:abliation_img}, we observe that introducing Danmaku leads to more compact intra-class clusters and clearer inter-class boundaries, demonstrating its effectiveness in guiding representation learning.

Finally, when all three components are jointly applied, the model achieves the best performance on both datasets. Compared with the base model, the accuracy improves by 5.52\% on FakeSV and 5.85\% on FakeTT. As shown in Fig.~\ref{fig:abliation_img} (DM-FEND), the feature distributions exhibit the most distinct separation between real and fake samples, with minimal overlap and well-structured clusters. This demonstrates that the three modules are highly complementary, and that modeling temporally aligned Danmaku as dynamic supervision is crucial for effective multimodal fake news detection. In section B of the supplementary document, we analyze the performance of replacing Danmaku with traditional comments in this framework.

\subsection{Case Study on \emph{Genda}}

\begin{figure}[ht]
    \centering
    \includegraphics[width=1\linewidth,height = 0.5\linewidth]{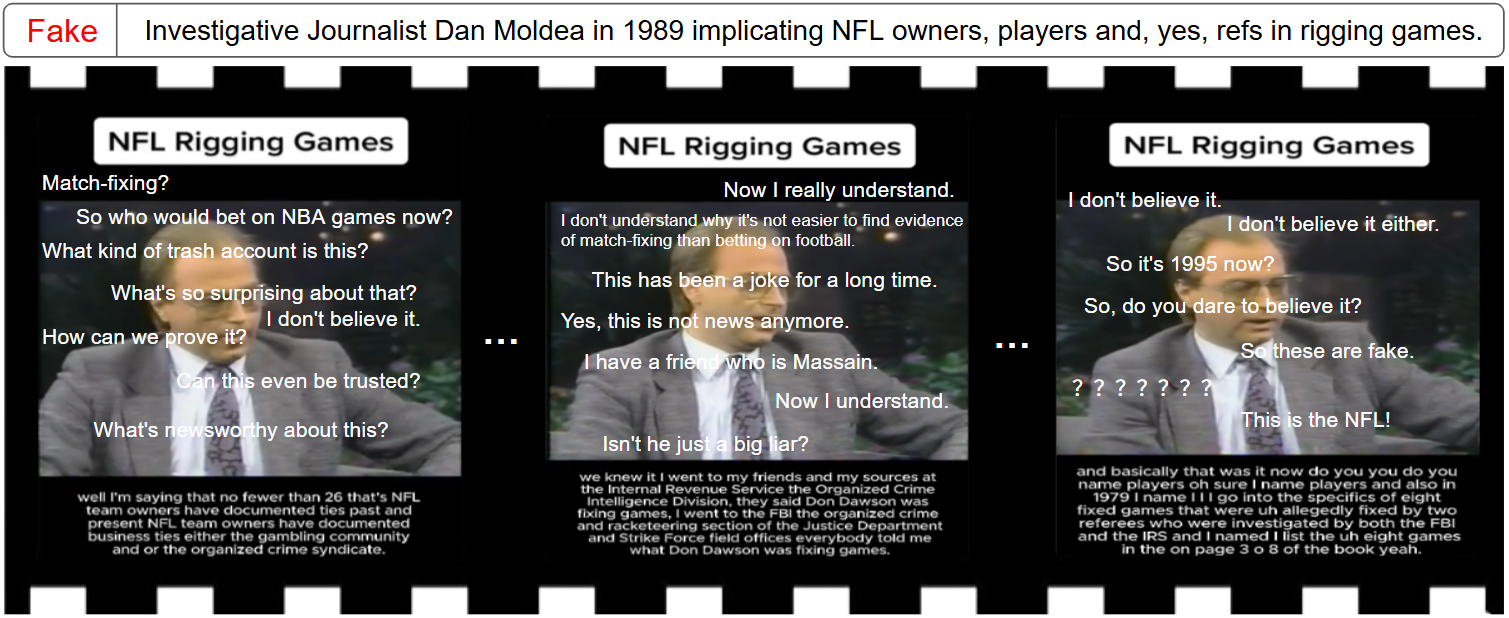}
    \caption{Case study of a short video with the video clip-level pseudo Danmaku generated using our proposed \emph{Genda} framework.}
    \label{fig:Case}
\end{figure}

To further illustrate the effectiveness of Danmaku in capturing fine-grained temporal signals, we present a case study in Fig.~\ref{fig:Case}.
From a temporal perspective, Danmaku is closely aligned with the corresponding video segments. In Clip 1, the video mainly introduces the event, while the Danmaku captures initial reactions such as surprise and mild skepticism (e.g., “Match-fixing?”). These comments match the visual content, where no clear contradiction has yet emerged.
In Clip 2, as the action unfolds, the Danmaku becomes more specific and directly responds to the scene. Comments such as “Yes, this is not news anymore.” and “Now I understand.” correspond closely to the visual evidence, showing that viewers react to concrete details rather than making generic judgments.
In Clip 3, when the critical text appears, the Danmaku strongly highlights deceptive signals. Statements such as “So it’s 1995 now? These are fake.” directly identify inconsistencies, suggesting that this segment contains key evidence of falsification. This shows that Danmaku can reveal the moments most indicative of misleading or manipulated content.

Overall, this case shows that the pseudo Danmaku generated by \emph{Genda} is not only temporally aligned with video content but also provides fine-grained cues that help localize potential fake signals. Such alignment enables the model to identify when and where inconsistencies occur, thereby improving the precision of fake news detection. We have also added failure cases in section C of the supplementary document. %, rather than relying solely on global video representations

\section{Conclusion}

This paper initially attempts to utilize Danmaku, which is known as bullet comments, that usually occur in short videos, for facilitating the detection of multimodal fake news. To bridge the gap between the cumulative latency of the Danmaku streams and the necessity of real-time fake news detection,  we introduce a novel temporal-aligned Danmaku generation framework, namely \textit{Genda}. Specifically,  \textit{Genda} simulates the dynamic process of user reactions during video playback by integrating a Danmaku Trigger and a Danmaku Generator, enabling the reconstruction of realistic interaction signals even in the absence of real-time user feedback. To make the generated Danmaku using \textit{Genda} good for usages, we further propose \textit{DM-FEND}, a Danmaku-guided temporal multimodal fake news detection model that supports fine-grained interactions among video, audio, text, and Danmaku at the clip-level of videos. Furthermore, \textit{DM-FEND} enhances unimodal representations by masking important regions and noise regions under the guidance of Danmaku, as well as strengthening cross-modal alignment through Danmaku-aware multimodal learning. The experimental results on two real-world short video datasets demonstrate that \textit{DM-FEND} outperforms various types of baselines, including unimodal detectors, LM-based benchmarks, and SOTA task-specific multimodal fake news detection approaches. In the future, we will incorporate contextual information of news, such as external knowledge or social propagation networks, to further improve the realism and reliability of Danmaku. %The ablation study further verifies the effectiveness and complementarity of different components embedded in \textit{DM-FEND}. In addition, the qualitative study shows that the generated Danmaku aligns well with real-world patterns and can accurately localize segment-level misleading signals.

\begin{acks}
This study is partially supported by the National Natural Science Foundation of China (62403412, 62273248), the Natural Science Foundation of the Higher Education Institutions of Jiangsu Province of China under grant 23KJB520040, the National Language Commission of China (ZDI145-71), and the Open Project Program of Key Laboratory of Knowledge Engineering with BigData (the Ministry of Education of China, NO.BigKEOpen2025-06).
\end{acks}

\bibliographystyle{ACM-Reference-Format}
\balance
\bibliography{sample-base}


\appendix

\end{document}